%% file: main.tex
\documentclass[letterpaper, 10 pt, conference]{ieeeconf}

\IEEEoverridecommandlockouts  

\usepackage{color}
\usepackage{array}

\usepackage{graphicx} 

\usepackage{mathptmx} 
\usepackage{times} 
\usepackage{amsmath} 
\usepackage{amssymb}  
\usepackage{xcolor}
\usepackage{bm}
\usepackage{soul}
\usepackage{rotating}
\usepackage{subfigure}
\usepackage{tabularx}
\usepackage{colortbl}
\usepackage{hhline}
\usepackage{multirow}
\usepackage{verbatim}
\usepackage{cite}
\usepackage{siunitx}
\usepackage[capitalise,noabbrev]{cleveref}
\usepackage{graphicx}
\usepackage{duckuments}
\usepackage[capitalise,noabbrev]{cleveref}

\usepackage[acronym]{glossaries}
\newacronym{hri}{HRI}{Human-Robot Interaction}
\newacronym{imu}{IMU}{Inertial Measurement Unit}
\newacronym{emg}{EMG}{Electromyography}
\newacronym{cnn}{CNN}{Convolutional Neural Network}
\newacronym{rnn}{RNN}{Recurrent Neural Network}
\newacronym{lstm}{LSTM}{Long Short-Term Memory}

\input{lib.tex}
\begin{document}

\setstcolor{red}

\title{\LARGE \bf
	Improving Tactile Gesture Recognition with Optical Flow
}


\author{Shaohong Zhong$^{1,*}$, Alessandro Albini$^{1,*}$,  Giammarco Caroleo$^{1}$ Giorgio Cannata$^{2}$ and Perla Maiolino$^{1}$
\thanks{$^{1}$ is with the Oxford Robotics Institute (ORI), University of Oxford, UK.}
\thanks{$^{2}$ is with the Department of Informatics, Bioengineering, Robotics and Systems Engineering (DIBRIS), University of Genoa, IT.}
\thanks{$^{*}$Equal contribution.}
\thanks{This work was supported by the SESTOSENSO project (HORIZON EU-
	ROPE Research and Innovation Actions under GA number 101070310).}
\thanks{We would like to acknowledge the use of the SCAN facility in carrying out this work.}
}

\maketitle


\input{abstract}
\input{Sections/intro}

\input{Sections/methodology}

\input{Sections/setup}

\input{Sections/results.tex}

\input{Sections/conclusion.tex}

{\small
	\bibliographystyle{IEEEtran}
	\bibliography{ref}
}

\end{document}

%% file: abstract.tex
\begin{abstract}
Tactile gesture recognition systems play a crucial role in \gls{hri} by enabling intuitive communication between humans and robots.
The literature mainly addresses this problem by applying machine learning techniques to classify sequences of tactile images encoding the pressure distribution generated when executing the gestures. However, some gestures can be hard to differentiate based on the information provided by tactile images alone. %

In this paper, we present a simple yet effective way to improve the accuracy of a gesture recognition classifier. 
Our approach focuses solely on processing the tactile images used as input by the classifier. In particular, we propose to explicitly highlight the dynamics of the contact in the tactile image by computing the dense optical flow. 
This additional information makes it easier to distinguish between gestures that produce similar tactile images but exhibit different contact dynamics.
%
We validate the proposed approach in a tactile gesture recognition task, showing that a classifier trained on tactile images augmented with optical flow information achieved a 9\% improvement in gesture classification accuracy compared to one trained on standard tactile images.
\end{abstract}

%% file: Sections/intro.tex
\section{Introduction}

\glsresetall

Research on \gls{hri} aims to provide smooth cooperation between humans and robots, allowing operators to interact with robots in the most natural way~\cite{gervasi2020conceptual}. In this respect, a significant body of literature studies techniques to send commands to robots using gestures~\cite{liu2018gesture,xia2019vision}. 
These gesture recognition systems predominantly rely on the recognition of motions of the human body or hands and are used to trigger specific robot actions.
Most of these systems are based on visual feedback obtained with the use of RGB or RGB-D cameras \cite{hasanuzzaman2004,ferng2009,sigalas2010,luo2012,qi2021}. Additionally, gesture recognition based on information acquired from \gls{imu} and/or \gls{emg} sensors has also been considered in the literature \cite{shin2014,carfi2018,singhvi2018}.

Beyond the classification of body motions, another possibility is to directly interact with robots through physical contacts~\cite{argall2010survey}. 
This type of communication is usually preferred when close \gls{hri} is required and where physical interaction is the most natural way to interact with the robot (e.g. to move the robot arm or to teach a movement \cite{albini2017}). In this respect, researchers have started taking advantage of tactile sensors to recognise human gestures applied to the robot \cite{tawil_2014}. 
%
%
In the literature, two main strategies have been proposed. The first is based on the processing of time series tactile data \cite{naya_1999,stiehl_2005,huisman_2013}, some of which also contains multimodal information \cite{kaboli_2015,koo_2008}. The second is based on the processing and classification of the pressure distribution, encoded as sequences of tactile images that describe how the contact shape evolves over a fixed time window \cite{tawil_2012,cirillo_2017,hughes_2018,salvato_2021}.

\begin{figure} [t!]
	\centering
	\includegraphics[width=\columnwidth]{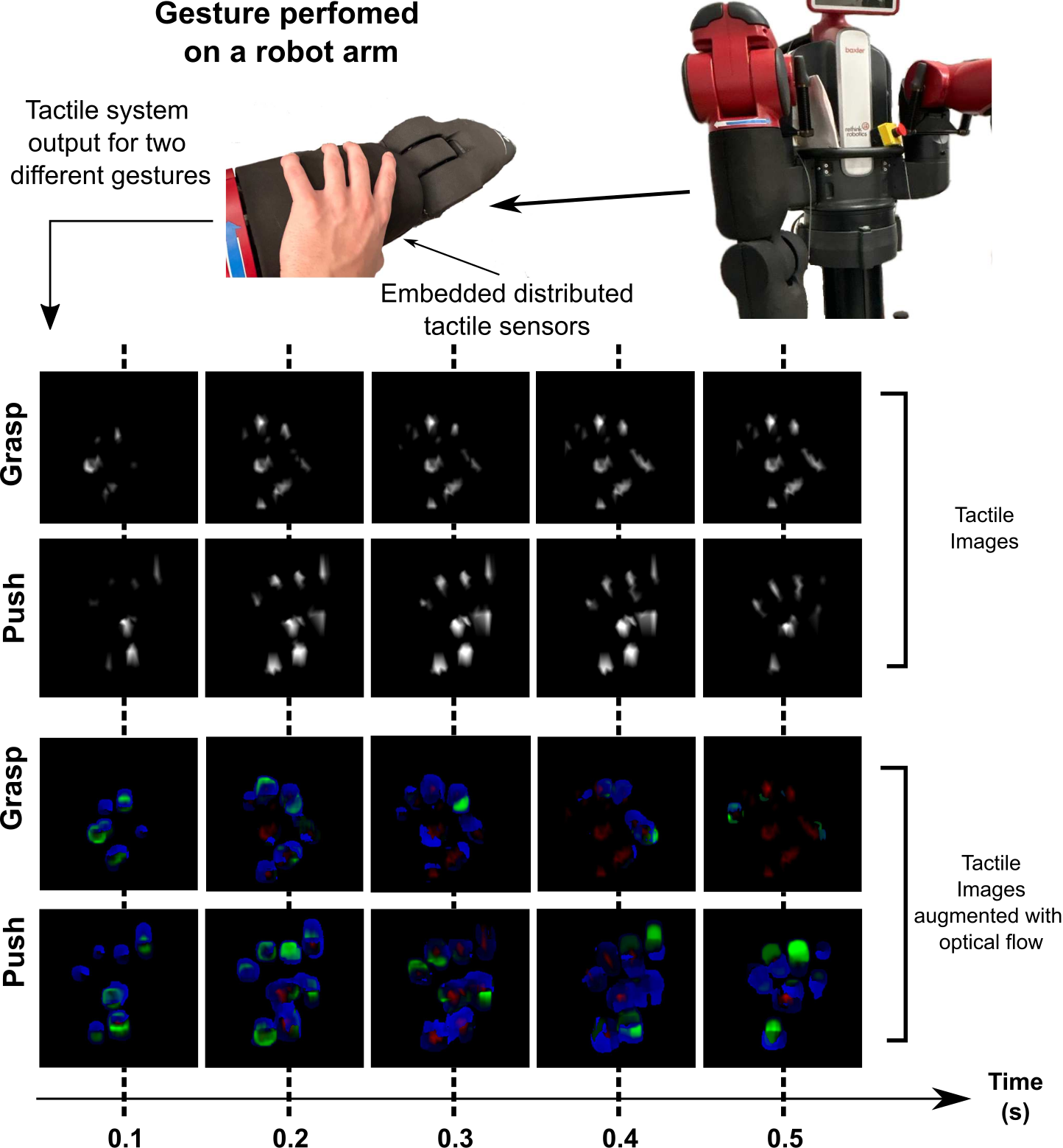}
	\caption{
 The gesture applied by the human to the robot arm generates a sequence of tactile images at each time instant (each \SI{0.1}{\second} as shown in the Figure). The contact shapes captured by the tactile images are similar between the two different gestures. However, with the proposed representation, which highlights the contact dynamics in the green and blue channels of the image, the differences become much clearer — after \SI{0.4}{\second}, the dynamic fades for the Grasp gesture, while is still present in the Push.
 }
	\label{fig:intro}
\end{figure} 

However, some gestures are not easily recognisable by analysing the contact shape alone. An example is given in \cref{fig:intro}, showing two sequences of tactile images corresponding to \textit{Grasp} and \textit{Push} actions performed by two different users. Each tactile image shows a greyscale representation of the pressure applied by the human hand at each time frame. As can be observed, both gestures involve the same parts of the hand (fingertips, palm and thumbs are present in both sequences), 
 making it hard to distinguish them solely on the basis of the contact shape.  
%
A possibility to make the gestures more distinguishable is to consider additional sensory inputs or the effects of shear forces \cite{choi_2022}. However, this requires additional complexity from the hardware point of view. Moreover, distributed sensors capable of capturing shear forces can hardly scale over large areas.

It must be noted that, although the contact shape may be similar among various gestures, the contact dynamics may be completely different. As an example, while a \textit{Grasp} is a quasi-static action, a \textit{Push} involves higher contact dynamics, i.e., the value of the pixels in the image significantly changes between two frames. This is not immediately visible from the sequences of grayscale tactile images in \cref{fig:intro}.

In this paper, we propose a simple and effective method to significantly improve the accuracy of a classifier
trained for tactile gesture recognition. This approach does not require any additional hardware and it is based on 
a processing of tactile images only. 
Similarly to what is commonly done in computer vision for camera-based gesture recognition~\cite{cheng2016,fang2007}, we exploit optical flow information to extract the dynamics of tactile gestures. Therefore, we propose to augment the tactile image by adding information on the variation of pixel intensities and displacement between consecutive time frames.
The proposed tactile image consists of 3 channels: the red channel represents the pressure distribution at the given time instant; the green and blue channels encode the optical flow in the form of a dense flow field \cite{horn1981}.

A sequence of augmented tactile images for \textit{Grasp} and \textit{Push} gestures can be seen in Figure \ref{fig:intro}. 
Compared with a standard tactile image, the contact dynamics are much more visible through augmentation, and we hypothesise that a classifier trained using augmented tactile images would obtain significantly better performance compared to the use of standard tactile images alone.
In this respect, we first collect a dataset of tactile gestures from a number of human users interacting with the robot. Then, we test our hypothesis by designing a classifier based on a \gls{cnn} and \gls{rnn} architecture \cite{sharma2021} trained on both types of input tactile image sequences.

The paper is structured as follows. \cref{sec:methodology} first describes the steps to obtain the proposed augmented tactile image from the sensors' raw data. Then, a description of the architecture of the gesture classifier follows. \cref{sec:setup} reports details on the experimental setup, the data collection procedure, and the training and evaluation details for the classifier. Results are presented and discussed in \cref{sec:result}. Conclusion follows.


%% file: Sections/methodology.tex
\section{Augmented Tactile Images for Gesture Recognition Tasks} \label{sec:methodology}

This section describes how the proposed tactile data representation is built, starting from raw tactile sensor measurements, and provides details on the network architecture used for gesture classification.

\subsection{Augmented Tactile Image} \label{sec:opt_flow}

\begin{figure} [t!]
	\centering
	\includegraphics[width=0.9\columnwidth]{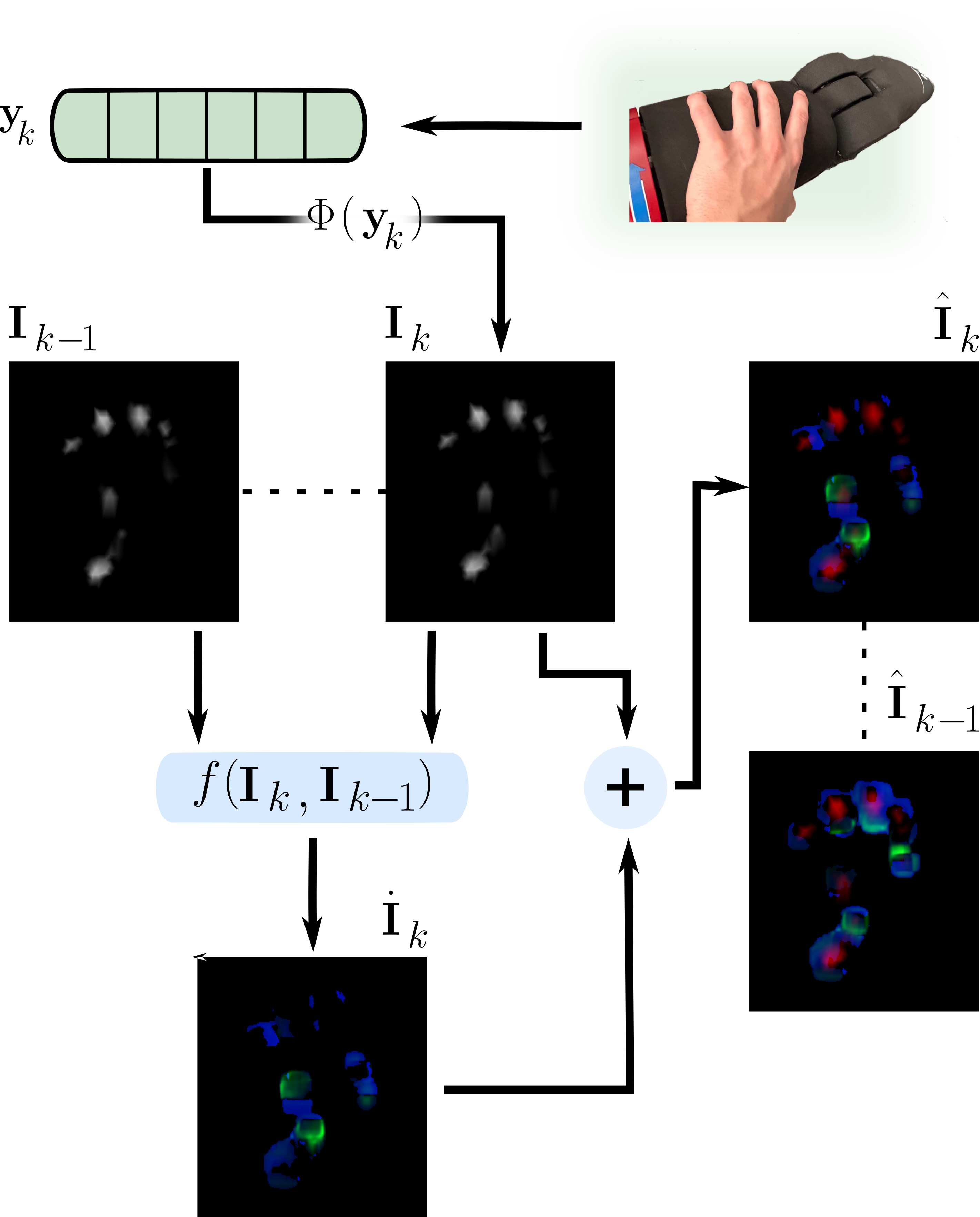}
	\caption{
 Processing steps required to augment a tactile image using dense flow information. As a gesture is performed on the robot, the distributed tactile sensors embedded in its body capture the pressure distribution, generating an array of measurements. The measurements are processed to create a 3-channel tactile image. The red channel represents the contact shape captured at the given time instant. The green and blue channels encode respectively the magnitude and direction of the optical flow computed between consecutive time frames.
 }
	\label{fig:opt_flow_build}
\end{figure} 

The complete processing pipeline used to create the tactile image augmented with dense optical flow information is shown in \cref{fig:opt_flow_build}.

It is assumed that the physical gesture is performed on an area of the robot equipped with distributed tactile sensors that can capture the physical interaction. In the following, we refer to large-area tactile sensing technologies composed of distributed transducers, namely \textit{taxels}, capable of providing information on the pressure applied over a certain region \cite{skin_survey,skin_6,skin_1,skin_2,skin_3}. At each sampling time instant $k$, the sensors produce a set of responses $\mathbf{y}_k = \lbrace y_{1_k}, y_{2_k}, \dots, y_{i_k}, \dots, y_{N_k} \rbrace$, where $y_{i_k} \in \Re$, $ i = \lbrace 0, 1, \dots, N \rbrace$, and $N$ is the total number of taxels. These responses contain raw measurements which are a function (typically non-linear) of the pressure applied on each taxel.

The array $\mathbf{y}_k$ can be converted to a tactile image through a process of resampling and interpolation of the taxels' spatial distribution \cite{pezzementi_2011, luo_2015}. In \cref{fig:opt_flow_build}, we refer to this transformation as $\Phi(\mathbf{y}_k)$. 
It must be noted that, if taxels are integrated over a non-planar robot body part (as in our experimental setup described in \cref{sec:setup}), the generation of the tactile image is still possible as described in \cite{albini2020}. 
The resulting tactile image is a 1-channel image, whose pixel values are related to the pressure applied on a specific area of the robot. As visible in \cref{fig:opt_flow_build}, the single image $\mathbf{I}_k$ captures the \textit{shape} of the hand in contact with the robot body when the user is performing the gesture.

The next step is to consider a sequence of two consecutive tactile images and compute the dynamic information. The optical flow is computed using the Gunnar Farnebäck algorithm \cite{gunnar2003}. The algorithm can be used to compute a dense optical flow, estimating the flow at each pixel of the image. Compared to other popular methods, such as the Lucas-Kanade algorithm \cite{lucas1981}, which assumes constant displacement of the pixels, it is computationally more expensive but more suitable for dealing with complex displacement patterns. Due to the relatively small resolution of tactile images, compared to high-resolution images acquired with cameras, the higher computational time is not critical for our application.

The dense optical flow $\dot{\mathbf{I}}_k = f(\mathbf{I}_k, \mathbf{I}_{k-1})$ is then computed between two consecutive frames, using the tactile image $\mathbf{I}_{k-1}$ generated in the previous sampling step. The first step is initialised with an image $\mathbf{I}_0$ whose values are null. $\dot{\mathbf{I}}_k$ consists of a 2-channel image — the first encodes the magnitude of the flow, while the second encodes its direction in polar coordinates.
The last operation consists of joining $\mathbf{I}_k$ and $\dot{\mathbf{I}}_k$ into a 3-channel image $\hat{\mathbf{I}}_k$, encoding, at time instant $k$, both the contact shape (red channel) and its dynamics, expressed as a dense flow — green for the magnitude and blue for the direction.

\subsection{Classification Architecture} 
\label{sec:net}

\begin{figure} [t!]
	\centering
	\includegraphics[width=\columnwidth]{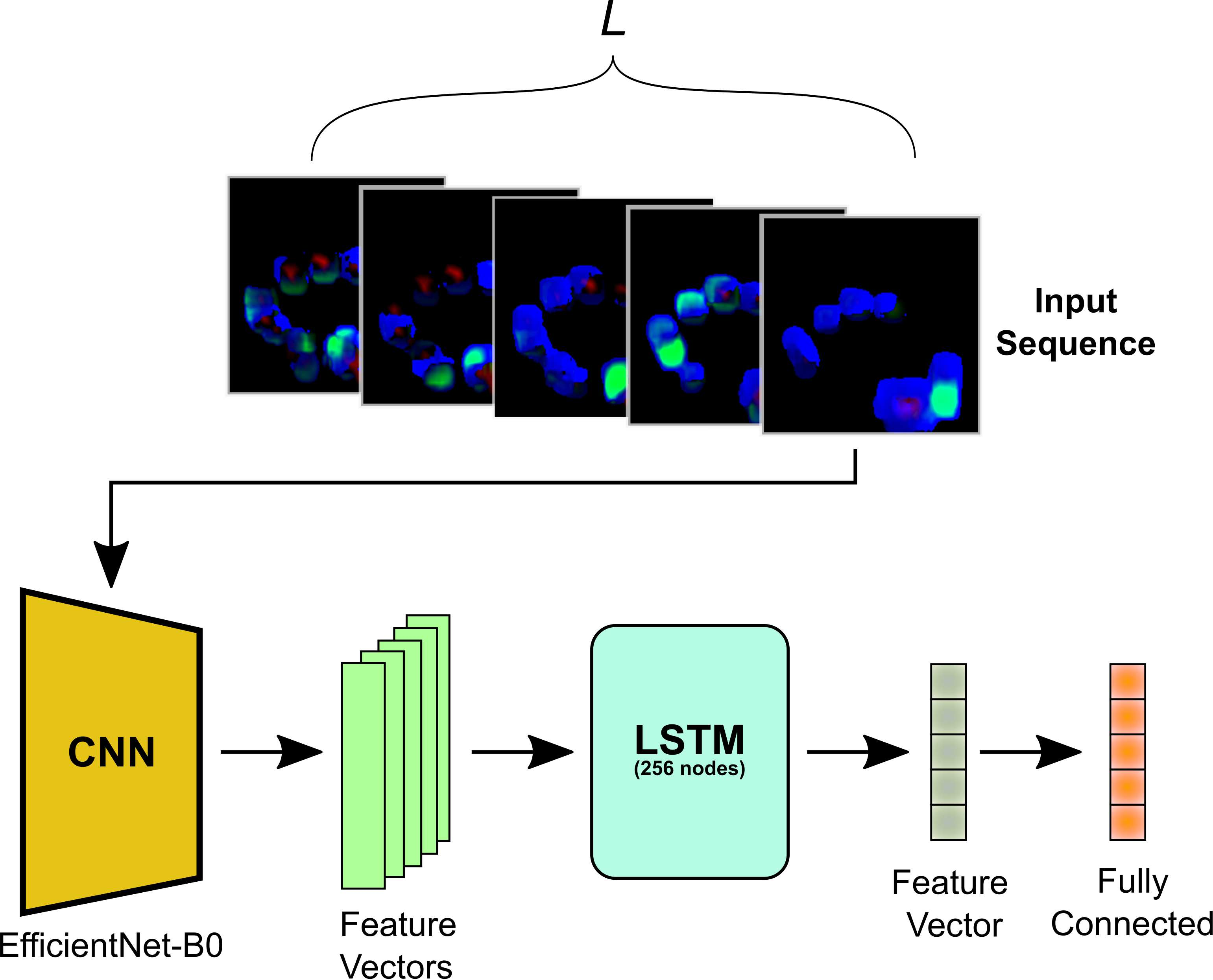}
	\caption{Classification model for gesture recognition based on a \gls{cnn}-\gls{lstm} architecture. The network takes a sequence of tactile images of length $L$ as input. }
	\label{fig:classifier}
\end{figure} 

To evaluate the effect of the proposed approach, we perform classification of the sequences of tactile images associated with different gestures.
Since a single data sample consists of a time series of tactile images, this resembles a video classification problem. Therefore, inspired by prior works on video classification~\cite{donahue2017long}, we leverage a combination of a \gls{cnn} and a \gls{rnn}.
For each input sequence, the \gls{cnn} is first used to extract spatial image features from each tactile image frame. Then, the sequence of image features is passed through the \gls{rnn} to capture the temporal dependencies between the frames. The output representation that captures both the spatial and temporal information is passed through a final layer of a fully-connected network to perform classification.

Specifically, for each sample, we perform simple preprocessing on individual image frames, including resizing and normalisation. Then, we use an \texttt{EfficientNet} pretrained on ImageNet to extract the image features from individual frames~\cite{tan2019efficient,deng2009imagenet}, and an \gls{lstm} network to capture the temporal information~\cite{hochreiter1997long}.
The choice of the \gls{cnn} and \gls{rnn} architecture is modular and can be adjusted depending on the size and type of the dataset. During training, the weights of all of the neural networks, including the pre-trained networks, are updated.

As visible in \cref{fig:classifier}, the input to the network is a sequence of images of fixed length $L$. The effect of changing the length $L$ on the classification accuracy is discussed in \cref{sec:setup,sec:result}.


%% file: Sections/setup.tex
\section{Experiments Description}
\label{sec:setup}

\subsection{Experimental Setup}

\begin{figure} [t!]
	\centering
	\includegraphics[width=\columnwidth]{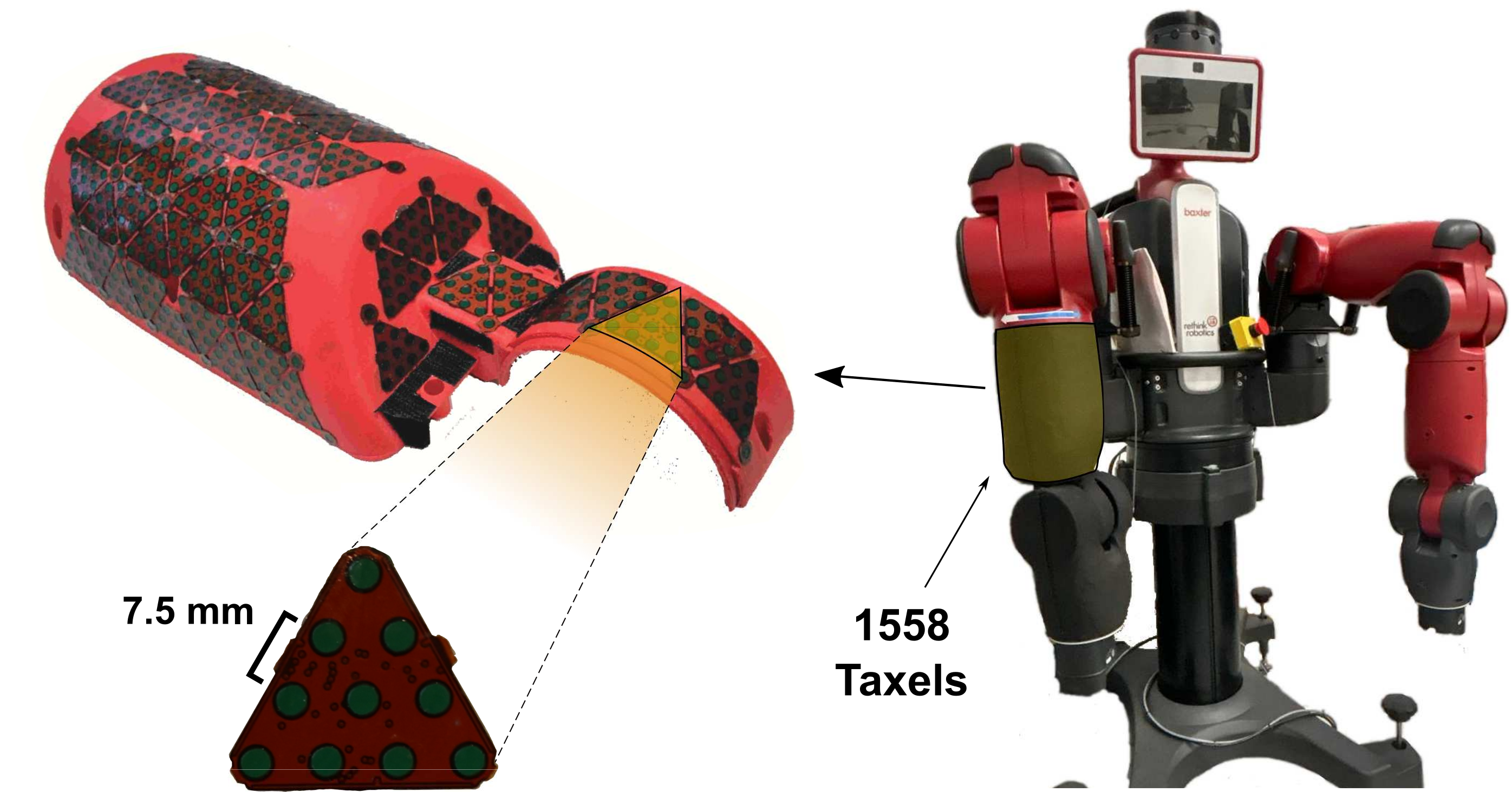}
	\caption{
Experimental setup: distributed tactile sensors are integrated into the forearm of the Baxter robot. The tactile system is composed of 1558 capacitive-based taxels, which acquire tactile measurements at \SI{10}{\hertz}. This platform is used to collect a dataset of 5 gestures from 38 people.
 }
	\label{fig:baxter}
\end{figure} 

The dataset has been collected on the platform shown in \cref{fig:baxter}, consisting of a Baxter robot equipped with 1558 distributed tactile sensors on the forearm. Sensors are covered by a conductive black fabric, and their placement (for the upper half of the forearm) can be seen in \cref{fig:baxter}. The tactile sensing technology, namely \textit{CySkin}, is an improved version of that presented in \cite{maiolino2011}. It is composed of triangular modules hosting up to 11 capacitive-based taxels, whose pitch is \SI{7.5}{\milli\meter}. Measurements are collected through a CAN bus at \SI{10}{\hertz} with 16-bit resolution.

\subsection{Tactile Dataset Collection}

\begin{figure*} [t!]
    \centering
	\subfigure[][] {
        \centering
        \label{fig:gesture_examples:original}
        \includegraphics[width=0.95\columnwidth]{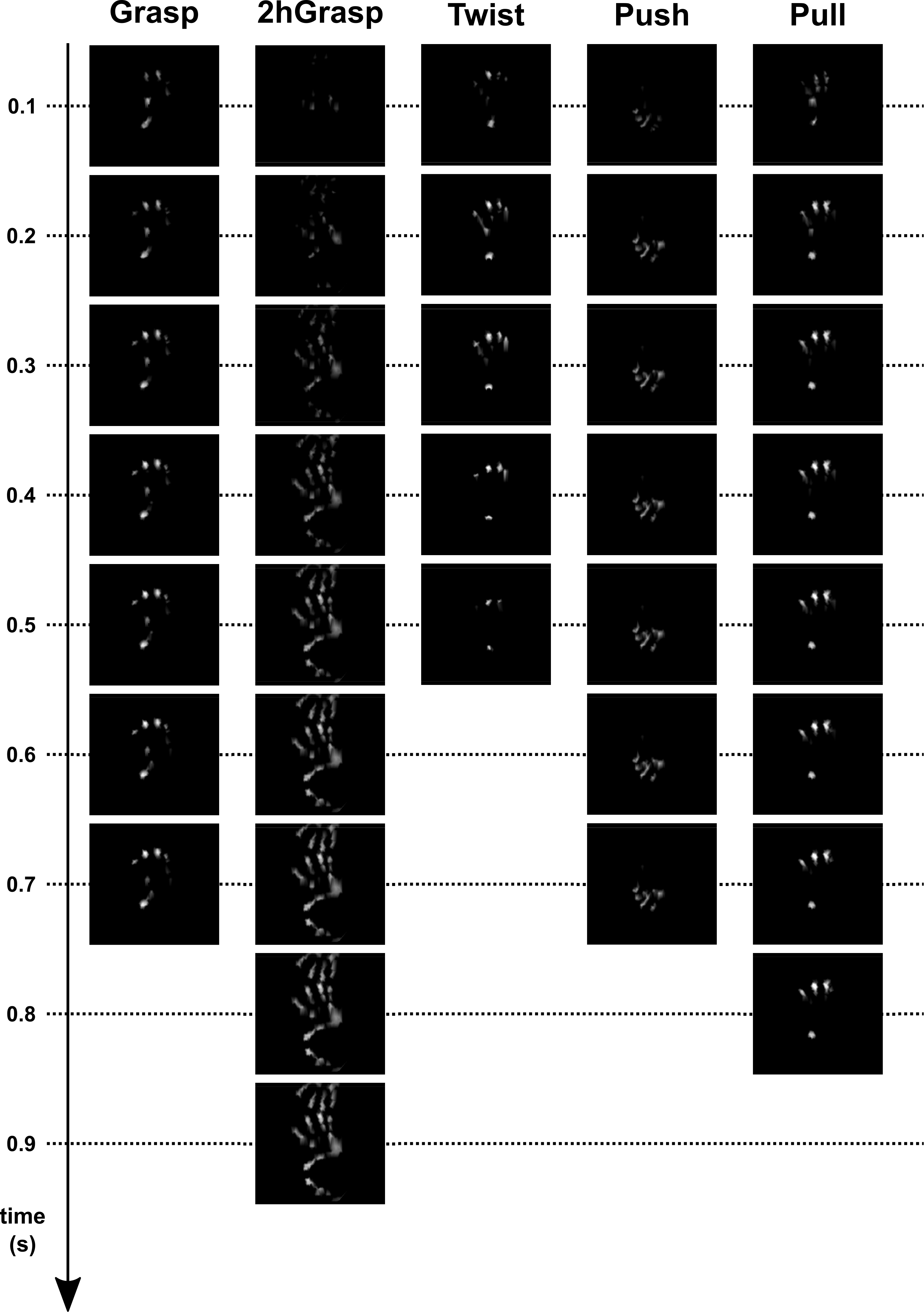}
        } \quad 
	\subfigure[][] {
        \centering
        \label{fig:gesture_examples:flow}
        \includegraphics[width=0.95\columnwidth]{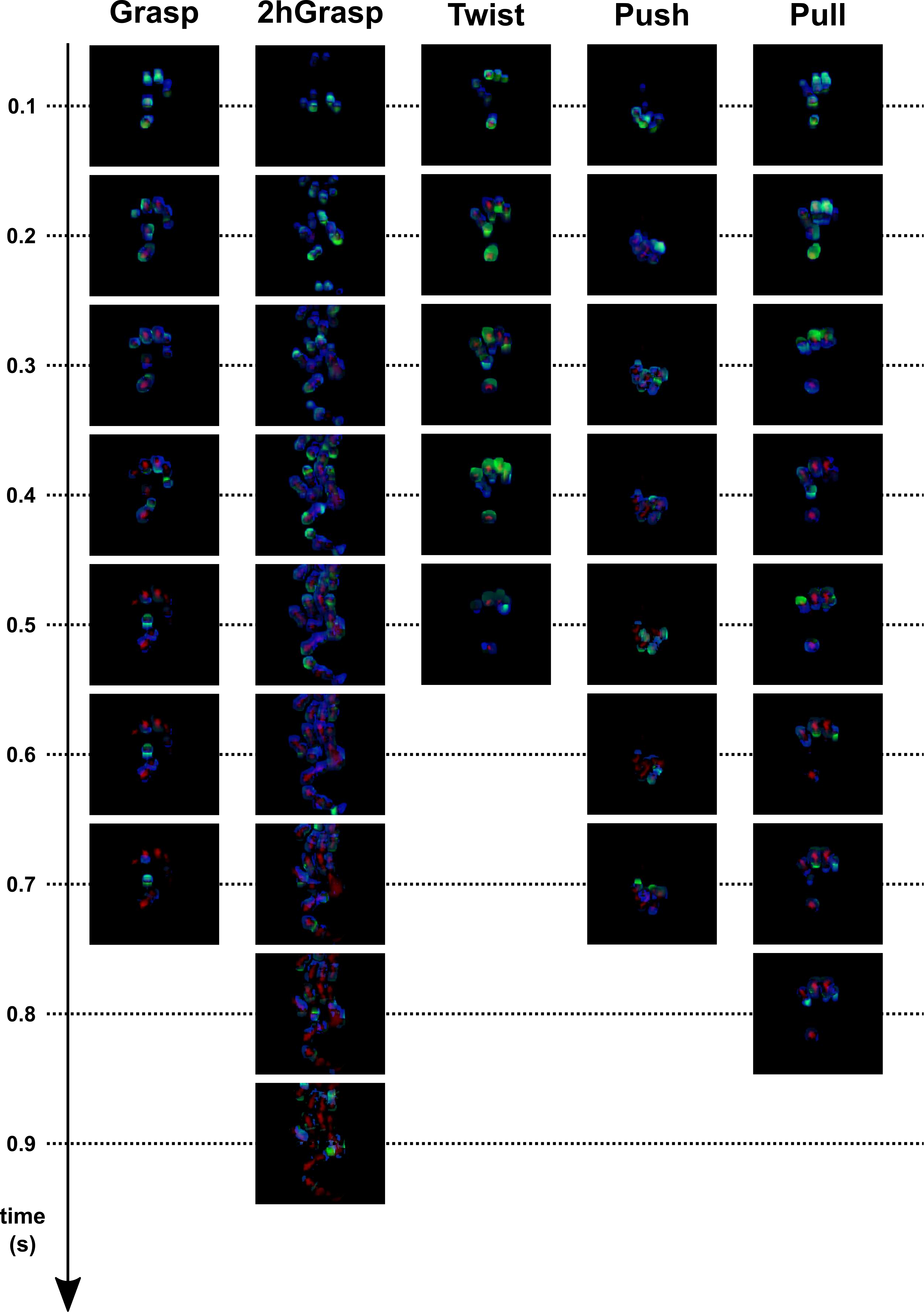}
        } 
	\caption{
 Tactile images corresponding to 5 different gestures performed by a single user. An image is generated at each sampling step of the tactile system (\SI{0.1}{\second}). As visible, using standard tactile images, Grasp, Twist and Pull look similar in this case. However, when using the proposed 3-channel encoding, their differences become more visible. (a) Sequences of standard 1-channel tactile images; (b) Sequences of tactile images embedding optical flow information.
 }
	\label{fig:gesture_examples}
\end{figure*} 

Users were asked to perform gestures on the robot's forearm, similar to those described in \cite{albini2020}:
\begin{itemize}
\item \textit{Grasp}: the user firmly grasps the robot forearm with one hand.
\item \textit{2hGrasp}: the user firmly grasps the robot forearm using both hands.
\item \textit{Twist}: the user grabs and twists the robot forearm along its rotational axis.
\item \textit{Push}: the user pushes the robot forearm away.
\item \textit{Pull}: the user grabs and pulls the robot's forearm.
\end{itemize}

It must be noted that when collecting these kinds of gesture datasets, data are affected by user variability \cite{naya_1999,choi_2022,albini2020}. Physical characteristics such as hand size or grasping strength may affect the way the user interacts with the robot. 
Since we wanted to let the users interact with the robot in the most natural way possible, participants were informed only by describing the gesture to be applied. No instructions were given about the force to be applied, the area of the forearm to be touched, or the length of the gesture. 
In order to take into account possible user variations and capture them in the dataset, we involved 38 users in the experiments\footnote{Users involved in the experiments signed an informed consent form.}. They differ in terms of gender (68.4\% Male, 31.6\% Female),
handedness (92.1\% Right, 7.9\% Left), age (21-34 years), weight (48-95 kg), and hand size (16-20 cm)\footnote{The hand size is measured from the base of the palm to the tip of the middle finger.}.
Compared to other studies on touch gesture recognition, this dataset has a number of gesture classes similar to \cite{tawil_2012,koo_2008,salvato_2021,cirillo_2017,hughes_2018}, while the number of people involved in the study is in line with \cite{salvato_2021,cirillo_2017} and much higher than \cite{kaboli_2015,naya_1999,huisman_2013,koo_2008,hughes_2018,tawil_2012}.

Each gesture was repeated 5 times in two different positions—standing in front of and on the side of the robot. Considering all 5 gestures, this led to 50 samples for each user, corresponding to a total of 1900 samples. During the whole experiment, the robot was commanded to maintain its pose. Each recorded sample consists of a sequence of tactile measurements collected every 0.1 seconds. The length of the gestures varies from a minimum of 0.5 s to a maximum of \SI{2.7}{\second}. Such a difference in length is due to the fact that after performing the gesture, users were still holding the robot for a while. In order to avoid the classifier being biased by the length of the gesture, we set the input size of the network to $L=5$, corresponding to 0.5 s, i.e., the shortest duration in our dataset. Furthermore, when testing the classifier, only the first $L$ tactile frames are considered. Additional analysis on the input length $L$ is reported in \cref{sec:result}.

From the raw measurements, two datasets of tactile images were generated. The first dataset is composed of 3-channel tactile images generated as described in \cref{fig:opt_flow_build}. The second dataset contains standard 1-channel tactile images and is needed as a comparison to analyze the advantage of including optical flow information in the input data to the classifier. The size of each image is 357 $\times$ 334. They have been created by resampling the distribution of the taxels as described in \cite{albini2020} with a spatial sampling step of \SI{1}{\milli\meter}. 

Finally, \cref{fig:gesture_examples:original} reports tactile images of the 5 gestures performed by a single user and sampled at 0.1 s\footnote{The contact shape in the 2hGrasp appears to be split due to the transformation of the 3D taxel distribution (wrapped around the forearm) into a 2D image \cite{albini2020}.}.
As visible from the contact shape, this user uses her/his whole hand to perform the \textit{2hGrasp} gesture, while mostly the fingers were involved in the \textit{Push} operation. Regarding \textit{Grasp}, \textit{Twist}, and \textit{Pull}, the contact shape is similar among the three gestures, since the participant mostly involved the fingertips in these three gestures. 
\cref{fig:gesture_examples:flow} shows the the same data processed as described in \cref{sec:opt_flow}. As shown, when superimposing the dense flow over the contact shape, \textit{Grasp}, \textit{Twist}, and \textit{Pull} become much more distinguishable. In particular, in the \textit{Grasp} case, the dynamic component quickly fades, and the static component of the pressure distribution becomes dominant at the fourth sample. In the \textit{Twist} action, the green channel (corresponding to the magnitude of the optical flow) is most dominant and is concentrated at the fingertips where the user is increasingly applying force to twist the robot forearm. Similarly, for the \textit{Pull} gesture, the dynamic part is much more evident than in the \textit{Grasp} and with a different pattern than the \textit{Twist}.
%
However, this is just an example -  as previously discussed, the way users interact with the robot may vary. 
For instance, the \textit{Push} gestures in \cref{fig:intro,fig:gesture_examples} are completely different - in \cref{fig:intro} fingertips, thumb and palm are involved, whereas only the fingers are visible in \cref{fig:gesture_examples}.

\subsection{Model Train and Test Details}

Both datasets were split into train and test. To ensure the validity of the average accuracy result, we randomly sample a fixed number of 30 samples ($\sim$10\%) from each gesture class to form the test dataset, and use the rest ($\sim$90\%) as the training dataset. Furthermore, we performed augmentation on the training dataset, whereby for each sequence longer than $L$ frames, we use a sliding window approach to generate additional sequences for training by slicing the full sequence into smaller sequences of $L$ consecutive frames, thereby generating more training samples. This results in $\sim$1800 samples for each gesture in the training dataset. 

The \gls{cnn}-\gls{lstm} network in \cref{fig:classifier} based on \texttt{EfficientNet-B0} (see \cref{sec:net}) was pretrained on ImageNet to extract the image features. We apply a dropout of 0.3 before passing the output of the \gls{lstm} network to a fully connected layer of 256 neurons for classification. For training, we use the standard cross entropy loss, and use stochastic gradient descent for gradient updates with a learning rate of 0.001. We train the models for 100 epochs.

For evaluation, we use three different seeds for each classification experiment and compare the results using the augmented 3-channel tactile images versus the original standard 1-channel tactile images. Furthermore, when analysing the effect of the length of the input sequence $L > 5$ in the second part of \cref{sec:result}, we pad the samples with insufficient frames with additional blank frames. 

%% file: Sections/results.tex
\section{Results}
\label{sec:result}

\begin{figure}[t]
	\centering
	\subfigure[][] 
	{\label{fig:cf_press}\includegraphics[width=0.85\columnwidth]{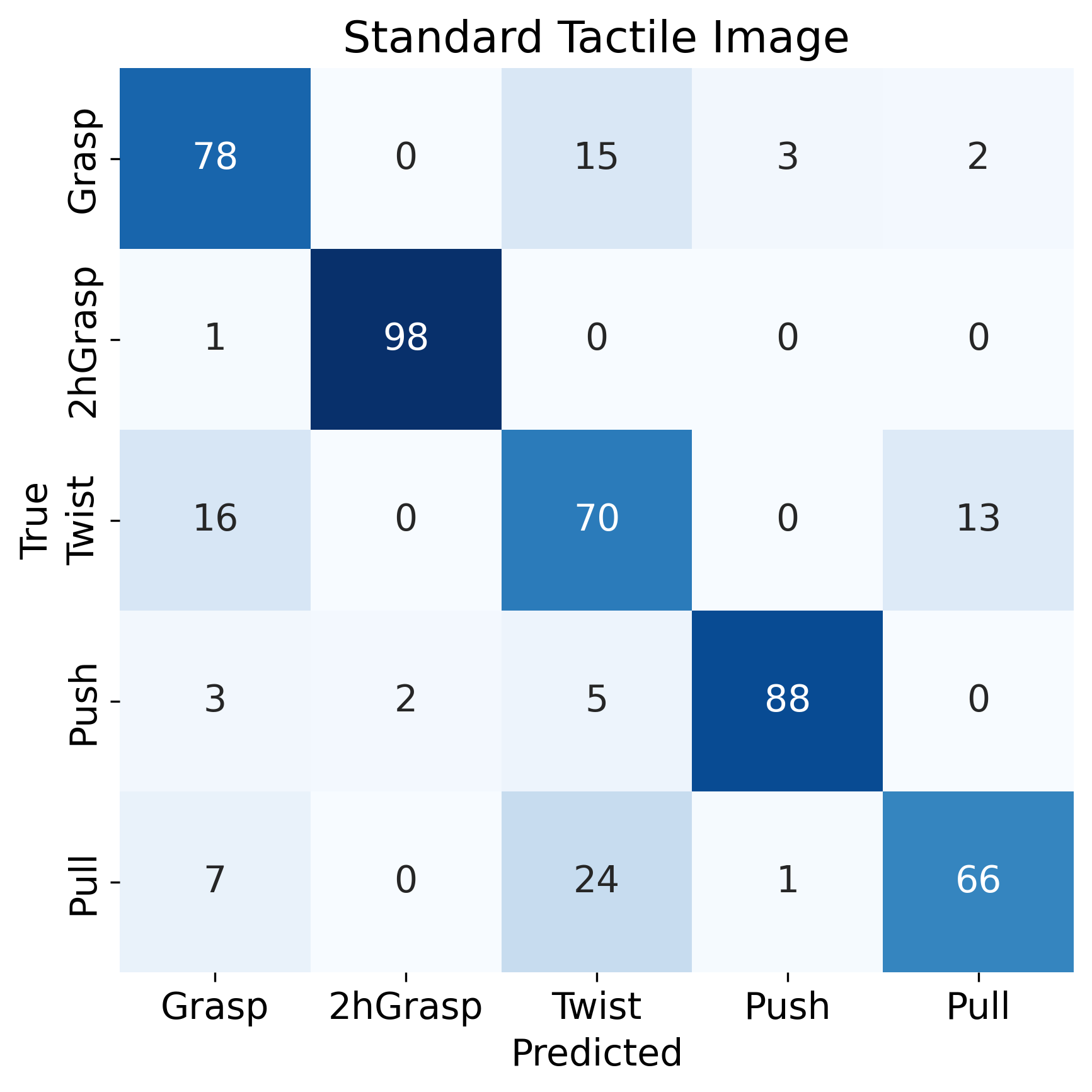}} 
	\subfigure[][]
	{\label{fig:cf_flow}\includegraphics[width=0.85\columnwidth]{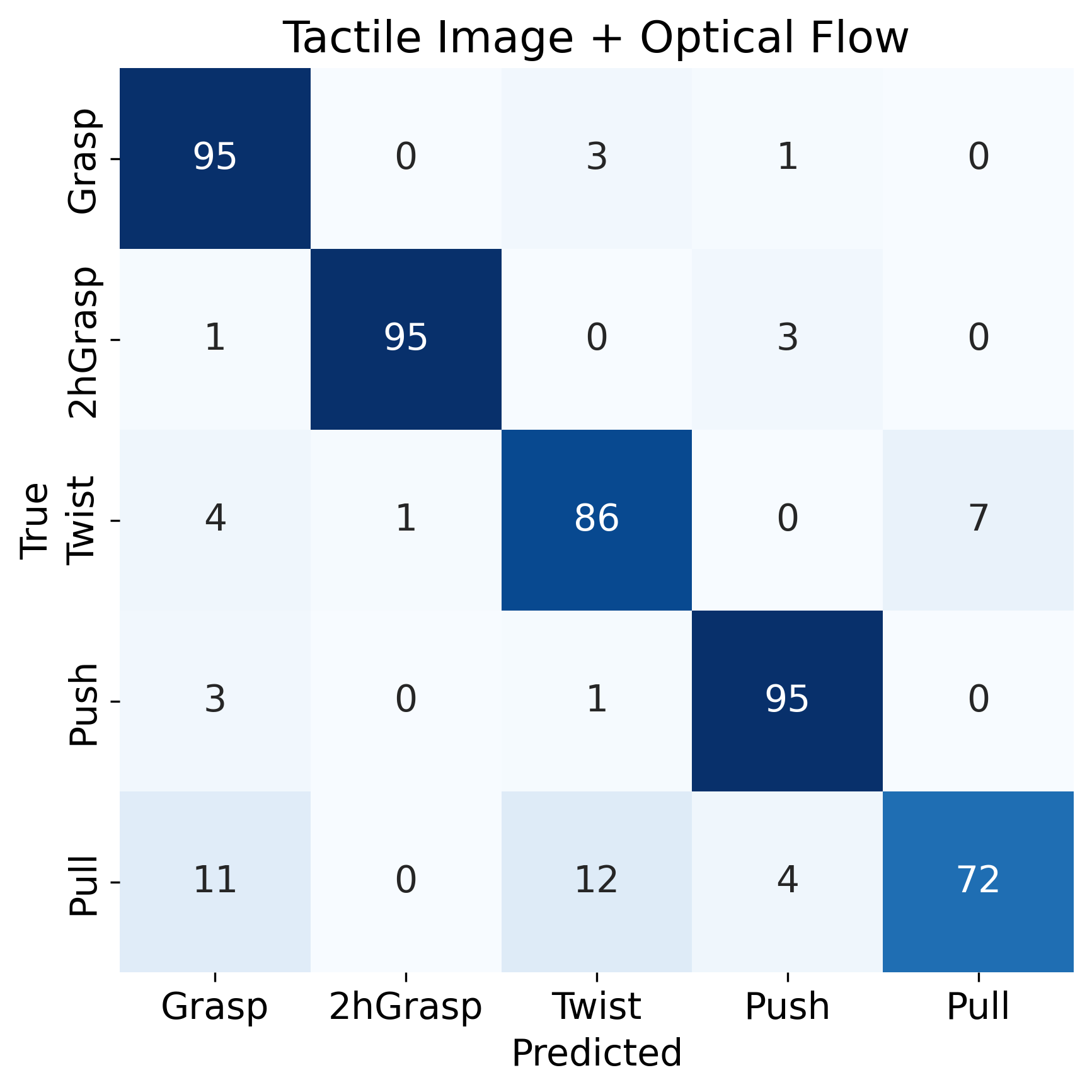}} 
	\caption{
		Normalised confusion matrices for the classification experiments. The results are averaged across three seeds. The proposed representation boosts the classification accuracy by about 9\%. (a) Confusion matrix obtained by training the model with standard tactile images - mean accuracy $80.7\pm0.4$\%. (b) Confusion matrix obtained by training the model with augmented tactile images - mean accuracy $89.1 \pm 0.2$\%.
  }
	\label{fig:conf_matrices}
\end{figure}

\cref{fig:conf_matrices} show the confusion matrices of the classifiers evaluated on the first $L = 5$ frames of sequences belonging to the test set. 

Using the original standard tactile images, the \gls{cnn}-\gls{lstm} classifier achieves a mean classification accuracy of $80.7\pm0.4$\% across the 5 gestures.
%
As shown in \cref{fig:cf_press}, the classifier trained on standard tactile images mainly confuses \textit{Grasp} with \textit{Twist} and \textit{Twist} with \textit{Pull}, while \textit{Push} is mainly confused with \textit{Grasp} and \textit{Twist}.

On the contrary, the proposed augmented tactile image allows the network to improve the classification accuracy by about 9\%, achieving $89.1 \pm 0.2$\%. This improvement is achieved solely by performing the simple preprocessing described in \cref{sec:opt_flow}.
From the confusion matrix in \cref{fig:cf_flow} it is clear that \textit{Grasp}, \textit{Twist}, \textit{Push}, and \textit{Pull} are much more distinguishable. The recognition of \textit{Grasp} improved by 17\%, reaching 95\%. Similarly, the accuracy for \textit{Twist} increased by 16\%. Compared to other gestures, \textit{Pull} only gains an increment of 6\% and is mostly confused with Twist. We argue that this is mainly due to the fact that both gestures primarily involve the fingertips. Although the dynamics are different for most users, a larger dataset may be required to properly train the model to distinguish between these two gestures.

\begin{figure} [t!]
	\centering
	\includegraphics[width=\columnwidth]{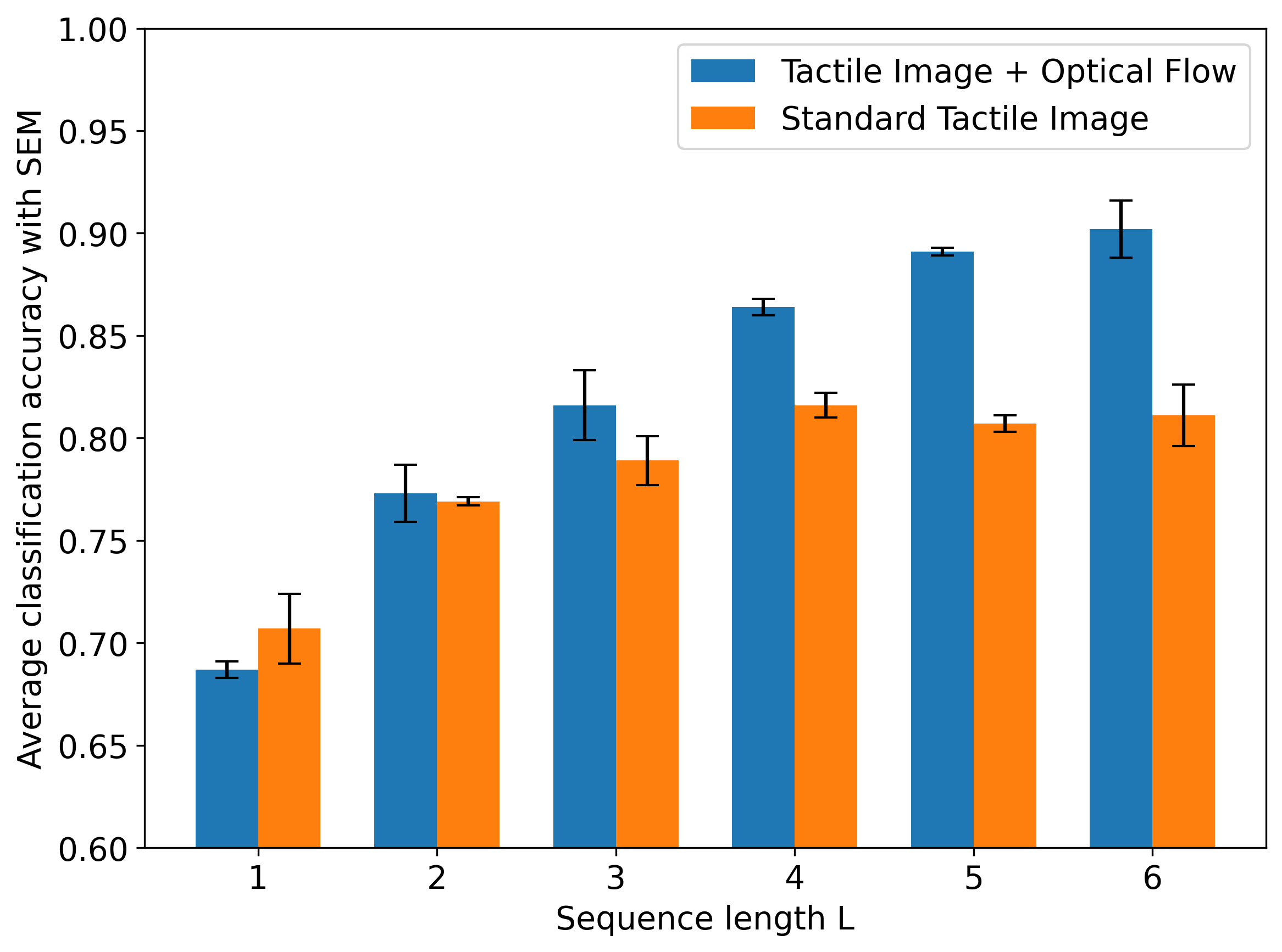}
	\caption{Trend of the mean classification accuracy by varying the input size $L$ from 1 to 6. }
	\label{fig:inp_length}
\end{figure} 

Finally, we also performed additional experiments by changing the input length $L$ and evaluating the effects on classification accuracy. In particular, smaller values of $L$ allow for classifying the gesture in a shorter time, thus reducing the delay between the user command and a possible action triggered by the robot in response to the specific gesture. On the contrary, larger values of $L$ increase the delay but allow for collecting more tactile information, possibly improving classification accuracy. In this respect, both classifiers have been retrained by considering $L = \lbrace 1, 2, 3, 4, 6 \rbrace$.

As shown in \cref{fig:inp_length}, in the case of tactile images augmented with optical flow, $L = 4$ frames already provide a reasonably good classification accuracy. With $L < 4$, the mean accuracy is below 85\%, with lower repeatability among the seeds for $L = 2$ and $L = 3$. For $L = 6$, as expected, the accuracy slightly improves to more than 90\%.
Regarding the model trained on standard tactile images, it is clear that performance is always lower than the model trained with augmented images. Furthermore, for $L \geq 3$, the accuracy reaches a plateau, and even for $L = 6$, it remains lower than what was obtained with the augmented tactile images using half of the input length.
%



%% file: Sections/conclusion.tex
\section{Conclusion}

In this paper, we analyse the effect of incorporating dynamic contact information into tactile data to improve the performance of gesture recognition tasks. Specifically, we propose a representation of tactile data as a 3-channel image, which integrates both the contact shape and its variation over consecutive frames through dense optical flow.

We validated this approach using a gesture classification task with a dataset collected from a large number of participants interacting with a robot. The results demonstrate that including optical flow information provides a simple yet effective solution to improve classification performance. Our experimental findings reveal a significant improvement in classification accuracy, highlighting the effectiveness of this method without necessitating additional hardware or complex machine learning architectures. 

Future works will be dedicated to extending the proposed representation by also including additional sensors commonly available on robots (such as torque sensors at the joints), and analysing how these can be exploited to further improve classification accuracy even while considering more complex gestures.